\documentclass[
]{ceurart}

\usepackage{listings}
\usepackage{xspace}
\usepackage[disable]{todonotes}
\usepackage{amsthm}
\newtheorem{definition}{Definition}

\newtheorem{elucidation}{Elucidation}

\newcommand{\gdc}{generically dependent continuant\xspace} 
\newcommand{\gdcs}{generically dependent continuants\xspace}

\newcommand{\gdcScary}{\textrm{GDC-Scary}\xspace}

\newcommand{\gdcAbb}{\textrm{GDC}\xspace}
\newcommand{\conc}{\textrm{concretizes}\xspace} 
\newcommand{\mani}{\textrm{carrier}\_\textrm{of}\xspace} 
\newcommand{\scary}{\textrm{Scary}\xspace} 
\newcommand{\inheres}{\textrm{inheres}\_\textrm{in}\xspace} 
\newcommand{\ornate}{\textrm{Ornate}\xspace} 
\newcommand{\gdcOrnate}{\textrm{GDC-Ornate}\xspace} 

\newcommand{\gdcDc}{\textrm{GDC-DC}\xspace}
\newcommand{\dc}{\textrm{DC}\xspace}
\begin{document}

\copyrightyear{2025}
\copyrightclause{Copyright © 2025 for this paper by its authors. Use permitted under Creative Commons License Attribution 4.0 International (CC BY 4.0).}

\conference{Proceedings of the Joint Ontology Workshops (JOWO) - Episode XI: The Sicilian Summer under the Etna, co-located with the 15th International Conference on Formal Ontology in Information Systems (FOIS 2025), September 8-9, 2025, Catania, Italy}

\title{Dispositions and Roles of Generically Dependent Entities}


\author[1]{Fabian Neuhaus}[%
]
\address[1]{Otto-von-Guericke-Universität Magdeburg, Germany
}
%
%


\begin{abstract}
 BFO 2020 does not support functions, dispositions, and roles of generically dependent continuants (like software or datasets). In this paper, we argue that this is a severe limitation, which prevents, for example, the adequate representation of the functions of computer models or the various roles of datasets during the execution of these models. We discuss the aspects of BFO 2020 that prevent the representation of realizable entities of generically dependent continuants. Two approaches to address the issue are presented: (a) the use of defined classes and (b) a proposal of changes that allow BFO to support functions, dispositions, and roles of generically dependent continuants. The latter also addresses limitations of BFO 2020 concerning the roles and dispositions of immaterial entities, particularly boundaries and sites.

\end{abstract}

\begin{keywords}
  Basic Formal Ontology \sep
  role \sep
  disposition \sep
 realizable entity \sep 
 quality \sep 
 generically dependent continuant \sep
 immaterial entities \sep 
 sites \sep 
 boundaries 
\end{keywords}

\maketitle

\section{Introduction}

Consider the following statements
\begin{enumerate}
	\item The software Protege has the function of  editing ontologies. 
	\item The fairy tale `Hansel and Gretel' has the disposition to scare children.  
	\item The Bundesgesetzblatt has the role of the public gazette of the Federal Republic of Germany, where laws are promulgated.  
\end{enumerate}

According to the  Basic Formal Ontology (BFO) these sentences are all false. 
BFO is a foundational ontology that is used by more than 600 different ontologies \cite{bfo-users} and has been published as ISO/IEC standard \cite{bfo-standard}. In BFO, software, stories, and publications are examples of generically dependent entities, and functions, dispositions, and roles are all kinds of realizable entities.  BFO claims that it is impossible for \gdcs to be the bearer of realizable entities. Consequently, the examples (1-3) all contain an ontological category mistake and, hence, are necessarily false.  

BFO's position seems to be at odds with ordinary discourse, in which we do not hesitate to attribute functions, dispositions, and roles to generically dependent continuants in the same way as we attribute them to material objects. 
For example, in the same way that civil engineers talk about the functional requirements of a bridge or an airport, software engineers talk about the functional requirements of software that they develop. 
 Nevertheless,  following BFO the results of the work of civil engineers bear functions (e.g., the function to provide an easy means to cross a river), while anything software engineers may create (e.g., an ontology editor) necessarily lacks functions. 
This is particularly implausible, since the function of many complex systems is based on the functions of their parts, which include both physical and software components. E.g., the brake of a car does not just include brake pads, rotors, and drums, but also a software component that controls the anti-lock braking system. If the software fails at realizing its function properly, the brakes may lock up and, thus, may fail to stop the car.  
Analogously, in some cases, we assign the same type of disposition to material entities and to \gdcs. For example,  in many countries, children do not have free access to alcoholic beverages or violent movies because it is widely assumed that both have the disposition to be harmful to children. However, according to BFO there is a major difference: while alcohol may have the disposition to harm children, violent movies do not. Because movies are \gdcs and, thus, have no dispositions at all.

These examples illustrate a limitation of BFO that its users encounter regularly, if they need to represent the realizable properties of \gdcs.
The main goal of this paper is to argue that BFO should be changed in a way that allows \gdcs to have functions, dispositions, and roles.  
 For this purpose, we will summarize briefly the relevant aspects of BFO in  Section~\ref{sec:relatedWork} and discuss why in BFO 2020  \gdcs may not be bearers of realizable entities.
In this context, we will argue that the definitions of roles and dispositions in BFO 20202 are too narrow, because they are not even applicable to immaterial entities like sites and boundaries.  
  Afterward, we will discuss a possible workaround that has been suggested by Barry Smith using defined classes (Section~\ref{sec:workaround}). As we will argue, the approach has the advantage that it does not require any changes to BFO, but it has severe drawbacks. For this reason, we propose an alternative approach, which  involves minimal changes to BFO that  enable realizable entities to inhere in immaterial entities and \gdcs (Section~\ref{sec:change}).

\section{Realizable Entities  in BFO}
\label{sec:relatedWork}
A review of the literature on functions, dispositions, and roles is way beyond the scope of this paper. However,  since the main purpose of this paper is to argue that BFO needs to be changed in a way that allows realizable entities to inhere in \gdcs, it is sufficient to focus on the literature of this kind of entity in the context of BFO.
There is a significant body of work on realizable entities in BFO \cite{
 arp_function_2008,
 rohl_representing_2011,
 rohl_why_2014,
 arp_building_2015,
 doi:10.3233/AO-160164,
 toyoshima2018formal,
 barton2018identity,
 toyoshima2021towards, 
 rabenberg2024grounding}. 
While the views represented in these papers diverge on many issues and details, they all agree on the major point that is relevant for this paper: \gdcs are not possible bearers of functions, dispositions or roles, because these kind of entities inhere in independent continuants\footnote{Typically, the authors only discuss only material objects as possible bearers of dispositions.}. Thus, in the remainder of this section, we will not discuss any of these papers in detail, but rather present a general overview, which is primarily based on the axioms and annotations in BFO 2020 \cite{bfo2020}. 
However, for the sake of brevity, we simplify some aspects of BFO that are not pertinent to our discussion.

\subsection{BFO Basics}
\label{sec:basics}
BFO distinguishes between three categories of entities that are significant for our purposes: independent continuants, generically dependent continuants, and specifically dependent continuants. Roughly speaking, independent continuants are entities that do not existentially depend on some other entity. The prototypical examples are material entities (which include objects like people and chairs), but the category of independent continuant also includes immaterial entities, which do not contain physical parts. These include sites, for example, a cave, the hold of a ship, or the Piazza San Marco in Venice. Another kind of immaterial entity is continuant fiat boundaries, which are distinguished by dimensionality. Examples include the boundary between two countries (a 1-dimensional boundary) or the Geographic North Pole (a 0-dimensional boundary). 

In contrast, specifically dependent entities are entities that are ontologically secondary in their existence and depend on another entity. E.g., the mass of the Eiffel tower or the function of the Hoover Dam \emph{specifically depend} on the Eiffel tower and the Hoover Dam, respectively. If a specifically dependent continuant $x$ specifically depends on an independent continuant $y$, we also say that $x$ \emph{inheres} in $y$ or that $y$ is the \emph{bearer} of $x$. 

Generically dependent continuants were a relatively late addition to BFO. They are entities that are neither independent nor dependent on a specific entity, because they are copiable entities. For example, the novel Moby Dick has been printed millions of times; each of these books (including ebooks and other representations) is a \emph{carrier} of the novel. The existence of Moby Dick does not depend on any of its carriers specifically, because even if one of its copies is destroyed, the novel continues to exist.
However, according to BFO, if all {carriers} of the novel cease to exist, Moby Dick ceases to exist. In this sense, Moby Dick is \emph{generically dependent} on its carriers.  
(Other examples of generically dependent continuants include computer programs, data sets, protein sequences, and engineering blueprints.) Thus, both generically dependent continuants and specifically dependent continuants existentially depend on independent continuants. Generically dependent continuants are closely linked to specifically dependent continuants. E.g., the arrangement of the ink on the pages of my volume of Moby Dick (a specifically dependent continuant) \emph{concretize} the novel Moby Dick (a generically dependent continuant). The novel could also be concretized in the electromagnetic arrangement that inheres in the chip of an e-book reader. In general, every carrier of a \gdc is the bearer of a specifically dependent continuant that concretizes the \gdc. Concretizations are often specifically dependent continuants, but may also be processes. E.g., a poem may be concretized by the shape of the ink on a page or by a performance on a stage.


%
%
%
%

In BFO 2020, functions are a kind of disposition. Dispositions and roles are kinds of realizable entities, which themselves are a kind of specifically dependent continuant.   
Realizable entities are introduced as follows: 
\begin{quote}
	A realizable entity is a specifically dependent continuant that inheres in some independent continuant which is not a spatial region and which is of a type some instances of which are realized in processes of a correlated type. \cite{bfo2020}	
\end{quote}

Both dispositions and roles existentially depend on a single bearer. The realization of a disposition occurs in virtue of the physical make-up of the bearer. Further, if a disposition of a bearer ceases to exist, then its bearer physically changes. For these reasons, dispositions are said to be \emph{internally grounded} in their bearer. 
In contrast, roles are \emph{externally grounded} in realizable entities. Hence, a bearer may lose a role without a change to its physical make-up. In the BFO literature, most examples for roles are social roles of physical objects, e.g., the role of being a student at a university or the role of a stone in marking a property boundary.  
However,  roles do not need to be social; consider, for example, the role of water bamboo as food for pandas. This realizable entity is not intrinsic to water bamboo, because if pandas became extinct, water bamboo would lose its role as panda food without a change in its physical make-up.

For two reasons, generically dependent continuants may have neither roles nor dispositions (and, consequently, neither functions) in BFO 2020. One is related to the inherence relationship, the other is related to the distinction between roles and dispositions.  

\subsection{Inherence and Dependency}
According to the quote above, realizable entities inhere in independent continuants. As mentioned above, inherence is a kind of specific dependence,  which may be defined as follows:  $x$ specifically depends on $y$ if and only if (a) $x$ is of a nature that, at any time $t$, $x$ cannot exist at $t$ if $y$ does not exist at $t$, (b) $x$ and $y$ do not share any parts and (c) $x$ is not a boundary of $y$. 

Following this definition of specific dependence, it seems that roles and dispositions (including functions) may depend on a generically dependent continuant. E.g., Protege's function cannot exist without Protege, and Protege's function is neither part of nor a boundary of Protege. However, BFO 2020 includes an additional assumption about the range of the `specifically depends on' relationship, which entails that no entity may specifically depend on a generically dependent continuant. This limitation is not just applicable to roles and dispositions, but also applies to qualities. Therefore, according to BFO 2020, generically dependent continuants may not have qualities. 

This is counterintuitive because we often ascribe the same qualities to independent continuants as to generically dependent continuants. However, according to BFO 2020, a sculpture may have the qualities of being baroque and ornate, but a poem can not. For the same reason, a person may have the quality of being scary, but, according to BFO 2020, there are no scary movies. 
The restriction also applies to relational qualities, for example, ownership. According to BFO 2020, you own your favourite T-Shirt, because there is an ownership quality that specifically depends on the T-Shirt and on you. 
However, since an ownership quality may not specifically depend on generically dependent continuants, it follows from BFO 2020 that nobody owns generically dependent entities, including novels, software, and patents. It seems unlikely that copyright and patent lawyers would agree with that aspect of BFO.  

The absence of the possibility of the qualities of generically dependent continuants is also problematic, because in BFO, the change of continuants is represented with the help of dependent continuants that specifically depend on continuants. For example, according to BFO, for each person (and independent continuant) there exists the mass of this person (a quality), which changes over time as the person gains or loses weight. Further, a person may gain and lose dispositions over time, e.g., many small children have the disposition to throw a tantrum when faced with adversity and lose this disposition when they grow up.  Generically dependent continuants change over time, too.  
For example, the United States Constitution was changed over time by amendments,  Microsoft Windows has become less buggy, and both grammar and vocabulary of the English language have changed significantly since Shakespeare's time. 
It is unclear how to represent these changes in BFO because the U.S. Constitution, MS Windows, and the English language are all generically dependent continuants and, according to BFO 2020, they lack qualities and dispositions, which could explain their change. 

\smallskip 
By assuming that qualities, roles, and dispositions may not specifically depend on generically dependent entities, BFO 2020 forces a distinction between independent continuants and generically dependent continuants that is at odds with how people typically talk about the world. Further, it raises difficulties for the users of BFO who need to represent the aspects of reality that, which -- to put it cautiously -- for all practical purposes seem to behave like qualities,  roles, and dispositions of generically dependent continuants. This would be justified if there were good ontological reasons for assuming that qualities, roles, and dispositions may not specifically depend on generically dependent entities. However, the literature on BFO does not provide a reason for this restriction.

\subsection{Distinguishing Between Roles and Dispositions}
In BFO 2020, roles are defined as optional, externally grounded, realizable entities. In contrast, dispositions are internally grounded realizable entities, which realizations (i) occur in virtue of their bearers' physical make-up and (ii) are causally linked to physical circumstances. Hence, the main criterion for distinguishing between roles and dispositions in BFO 2020 is directly linked to the physical make-up of their bearers. 
If one were to apply this criterion to generically dependent continuants, one would face the issue that they are not material entities. Thus, they do not have a physical make-up that may be changed, and, hence, the criterion for distinguishing between roles and dispositions is not applicable. 

This would be a good argument against introducing roles and dispositions of generically dependent continuants, if it were not the case that the criterion does not even work for all independent continuants in BFO. 
E.g., the Mason–Dixon Line is a continuant fiat boundary, which has the role to demarcate the four U.S. states\footnote{An annotation of Role in the bfo-core.owl file of BFO 2020  includes the example of a boundary that has the role of demarcating two neighboring administrative territories.}; and the Geographic North Pole has a role as a reference point for navigation and mapping. 
 Further, a cave may have the disposition to serve as shelter or the disposition to collapse. 
The Mason-Dixon Line, the Geographic North Pole, and caves are all examples of immaterial entities. Thus, some immaterial entities can be bearers of roles and dispositions. However, since immaterial entities have no physical make-up that may be changed, the criterion for distinguishing between roles and dispositions in BFO 2020 does not apply to them.\footnote{One could argue that the criterion is applicable, but that the lack of physical make-up of immaterial entities just entails that immaterial entities cannot have dispositions but only roles (and possibly other realizable entities that are not categorized in BFO). However, this would have the consequence that the same type of realizable entities are classified as dispositions of material entities, but not as dispositions of immaterial entities. E.g., the \emph{Propensity to collapse during an earthquake of magnitude 5}  of a building is a typical example of a disposition of a building, since the propensity is the result of the way the building was designed and the physical properties of its materials. But -- if we were to follow this argument --  the \emph{Propensity to collapse during an earthquake of magnitude 5} of a cave would not be classified as a disposition, since a cave is an immaterial entity in BFO. This seems arbitrary.  As we discuss in Section~\ref{sec:change}, in our opinion, this propensity is a disposition of the cave, because it is the result of the physical make-up of the rock that surrounds the cave.} Therefore, a change of a physical make-up as the criterion for distinguishing between roles and dispositions is too narrow and needs to be revised. In Section~\ref{sec:change} we propose definitions of role and disposition, which are applicable to material entities, immaterial entities, and generically dependent continuants.

\medskip 
To summarize, one reason why in BFO 2020 roles and dispositions do not inhere in generically dependent continuants is because (a) inherence is defined as a subtype of specific dependence and (b) specific dependence is defined in BFO 2020 with an extra condition that prevents qualities, roles, and dispositions from specifically depending on generically dependent continuants. By removing this extra clause, the definition of specific dependence would remain substantially unchanged, just a little bit less cluttered. 
The other reason is that the criterion for distinguishing between roles and dispositions in BFO 2020 depends on the physical make-up of their bearers. However, since this criterion does not fit immaterial entities (like boundaries and sites), the criterion is not really suitable for BFO anyway.  

\section{Approach A: Realizable Entities as Defined Classes?} 
\label{sec:workaround}
In a personal conversation, Barry Smith suggested that examples which seem to involve dependent entities (like qualities, roles, and dispositions) that inhere in generically dependent continuants should be analyzed in a way that the bearers of the concretizations of the generically dependent continuants are the bearers of the dispositions. 
Hence, the claim that   `Hansel and Gretel' has the disposition to scare children is not really about the fairy tale (i.e., the generically dependent continuant), but about all independent continuants (e.g., books or computer chips) that bear a specifically dependent continuant that concretize the `Hansel and Gretel' tale. 
More specifically, Barry Smith suggested that this approach may be implemented with the help of defined classes. He did not provide any details on how to implement this idea,  but in the following, a possible approach is presented.

As discussed in Section~\ref{sec:basics},  \emph{carrier\_of} is a relation that holds between an independent continuant and a \gdc if concretization of the \gdc inheres in the independent continuant. Generic dependence is the inverse of the carrier relationship.\footnote{For the sake of readability,  universal quantifiers at the start of the formulas are omitted. Further, we omit temporal arguments of the relationships, since the temporal dimension is not important for this analysis.  }   
\begin{definition}
	$
	\mani(x,y) \leftrightarrow  \exists z (\inheres(z, x) \wedge \conc(z, y))  	
	$
\end{definition}

Let's assume a story has the property of being scary if (and only if)  all of its carriers are scary. In this case, we may introduce a new defined class \gdcScary in the following way, where $\scary$ is a kind of disposition. 
\begin{definition}
\  \newline
$
\gdcScary(x) \leftrightarrow ( \gdcAbb(x) \wedge \forall y (  \mani(y,x)  \rightarrow \exists z (\scary(z) \wedge  \inheres(z,y)  ))))	
$
\label{def:scary}	
\end{definition}
In this case, \gdcScary is an example of a defined class, which we may call  `GDC-disposition'. 
However, this approach does not assume the existence of any particulars of type GDC-disposition. Rather, the defined class is instantiated by the generically dependent continuants. 
From a logical point of view, this is called a definitional extension of the original theory. All models that satisfy the original ontology may be trivially extended to satisfy the ontology that includes the additional definition. Thus, it does not change the formal semantics of the ontology. Any occurrence of \gdcScary in an ontology may be eliminated from the ontology at any time by replacing the predicate by its definition.  
 Hence, adding  \gdcScary to the ontology does not change the ontological commitments of the ontology. \gdcScary is just  fa\c{}on de parler, which only seems to involve dispositions of \gdcs, but is just a convenient way of talking  about the dispositions of its carriers. 

The same approach expressed in Definition~\ref{def:scary} may also be applied not just to other dispositions, but to any kind of dependent continuant (i.e., roles and qualities) by following the pattern in  Definition~\ref{def:pattern}, $\dc$ is a type of dependent continuants (e.g., the disposition of being scary or the role of being a sacred text) and $\gdcDc$ is the corresponding defined class for \gdcs.

\begin{definition}
$
\gdcDc(x) \leftrightarrow ( \gdcAbb(x) \wedge \forall y (  \mani(y,x)  \rightarrow \exists z (\dc(z) \wedge  \inheres(z,y)  ))))	
$
\label{def:pattern}
\end{definition}
For example, assume that the quality type \ornate is instantiated by the qualities of being ornate of various material entities like sculptures. Let's further assume that it is also instantiated by carriers of generically defined continuants (e.g., by a print of a poem in a given book). In this case, we may define \gdcOrnate as the class of \gdcs, which are carried only by ornate material entities (see Definition~\ref{def:ornate}).  
\begin{definition}
\  \newline
$
 \gdcOrnate(x) \leftrightarrow ( \gdcAbb(x) \wedge \forall y (  \mani(y,x)  \rightarrow \exists z (\ornate(z) \wedge  \inheres(z,y)  ))))
 $
 \label{def:ornate}
\end{definition}
While this approach may look promising at first, it has several major drawbacks.
First, in BFO, an instance of a class of dispositions (like soluble) is not a material entity,  but a disposition particular that inheres in the material entity. However, according to Definition~\ref{def:scary}  an instance of the class  \gdcScary is not some GDC-disposition particular, but the \gdc itself. This asymmetry is going to be quite confusing to users, which will lead to mistakes. (The same argument applies to other instantiations of the pattern in Definition~\ref{def:pattern}).

Second, as we discussed above, \gdcs may change.  In BFO, change of material entities is represented primarily by change of qualities and disposition particulars. For example,  my mass changes as I gain and lose weight, and my hay fever (i.e., the disposition of my immune system to overreact to allergens in the air) may weaken over time. Analogously, it seems that a change of \gdcs is related to a change of their quality and disposition particulars.  E.g., the length of the U.S. Constitution increased, and the disposition of Microsoft Windows to crash decreased over time. However, the main purpose of introducing defined classes following the pattern in  Definition~\ref{def:pattern} is to avoid the introduction of dependent particulars that inhere in \gdcs{}. Thus, this approach cannot represent the change of \gdcs{}.

Third, dispositions of material entities are usually closely connected to their qualities.\footnote{The relationship between dispositions and qualities is not explicitly represented in BFO 2020, but discussed in detail in  \cite{rohl_representing_2011, toyoshima2018formal, toyoshima2021towards}.  } 
 E.g., the disposition of a key to open a specific lock is based on its shape quality, and a significant change of its shape will cause the key to lose its disposition.   
For the same reason,  material entities that differ in their qualities typically differ in their dispositions. However, if a fairy tale is scary, the material base of its concretization does not matter. You may concretize `Hansel and Gretel' with the help of ink on paper, in braille, by electrical charges in computer chips, by punch cards, or in stone tablets. The physical qualities of these carriers are completely different. Thus, it seems implausible to claim that their physical make-up somehow leads to the same disposition, namely, the disposition to scare children.  
	
Finally, the whole approach is based on the assumption that all examples where dependent continuants (seem to) inhere in a \gdc may be explained by inherence relations between dependent continuants and the material entities that carry the \gdc.  
However, this assumption is problematic, especially for relational qualities like \emph{ownership}. E.g., if we follow this approach, the statement \emph{The Coca-Cola company owns the Coca-Cola logo} is just a short way for asserting the following: every material entity that carries the Coca-Cola logo is owned by the Coca-Cola company. However, this is not true.  If a customer buys a bottle of Coke, the customer owns this bottle, including the paint that carries the logo. There is no ownership relationship between the paint on the bottle and the Coca-Cola company, in spite of the fact that the company owns its logo. This example illustrates that, at least in some cases, a relational quality that inheres in an entity and a \gdc cannot be reduced to relational qualities that inhere in the entity and the carriers of the \gdc.

\medskip
For these reasons, the approach of using defined classes to represent realizable entities of \gdcs is not convincing. Instead, a solution requires some minor modifications to BFO. 
\section{Approach B: Changing BFO} 
\label{sec:change}
This section presents a possible approach that would enable BFO to support roles and dispositions of  \gdcs{}. Since the BFO developers tend to be conservative and, thus, like to avoid radical changes, the proposal is intended to minimize the necessary change. 
Hence, the following definitions and elucidations are based on BFO 2020 and only propose minor tweaks (highlighted in bold font). 

The definitions of specific dependence and inherence are modified by including \gdcs in the range of the relationships.  
\begin{definition}[Specific dependence]
	 $x$ specifically depends on $y$ if and only if 
	 (a) $x$ is a specifically dependent continuant, 
	 (b) $y$ is a specifically dependent continuant  \textbf{or a \gdc{}} or an independent continuant that is not a spatial region,
	 (c) $x$ is of a nature that, at any time $t$, $x$ cannot exist at $t$ if $y$ does not exist at $t$, 
	 (d) $x$ and $y$ do not share any parts and 
	 (e) $x$ is not a boundary of $y$. 
\end{definition}

\begin{definition}[Inherence]
	x inheres in y if and only if (a) x is a specifically dependent continuant and (b) \textbf{y is a \gdc or} y is an independent continuant that is not a spatial region and (c) x specifically depends on y. 
\end{definition}

The elucidation of realizable entities in BFO 2020 asserts that they are particulars of a type that is instantiated by some instances, which are realized in a corresponding process. The elucidation also asserts that these particulars inhere in independent continuants. This may be generalized to \gdcs{}.  
 
\begin{elucidation}[Realizable entity]
A realizable entity is a specifically dependent continuant that (a) inheres in some independent continuant (which is not a spatial region) \textbf{or inheres in a \gdc} and that (b) is of a type some instances of which are realized in processes of a correlated type. 
\end{elucidation}

The realization of a realizable entity of a \gdcs{} necessarily involves a concretization of the \gdc{}, either as part of the process or as a dependent continuant that inheres in a carrier of the   \gdc{}.
 E.g., the disposition of stories to be scary is realized in processes that involve frightened listeners. Each of these processes involves a concretization of the scary story (e.g., by being printed in a book or by telling the story). 

\begin{elucidation}
	\textbf{If a process $p$ is a realization of a realizable entity that inheres in a \gdc $x$, then either (a) a concretization of $x$ is part of $p$ or (b) some participant of $p$ is a carrier of $x$.}
\end{elucidation}

The notions of disposition and role need to be generalized in BFO in order to support entities without a physical make-up. For this purpose, we also consider parts and their arrangement, as well as retainers of immaterial entities.  A retainer of site is a material entity in relation to which the site is defined (\cite{arp_building_2015},  pp. 112-113). For example, your office is a site, and its retainer is the aggregation of its floor, walls, and ceiling. In \cite{arp_building_2015} the term `retainer' is only introduced in the context of sites.
 However, since a continuant fiat boundary is also a kind of immaterial entity whose location is determined in relation to some material entity,  we suggest to generalize its use for both sites and boundaries:   a material object $x$ is a retainer of a site or a continuant fiat boundary $y$ if the the location of $y$  is determined in relation to  $x$. 

Considering the mereotopology and the retainers of entities enables us to generalize the definition of Role and Disposition in a way that covers sites, and continuant fiat boundaries, and \gdcs. 

%
%

\begin{definition}[Disposition]
	x is a disposition if and only if (a) x is a realizable entity that inheres in some bearer y and (b) x ceasing to exist requires a change of  \textbf{physical qualities of y or the physical qualities of the retainer of y or the mereotopology of y}. 
\end{definition}

\begin{definition}[Role]
	x is a role if and only if (a) x is a realizable entity that inheres in some bearer y and (b) x may cease to exist without a change of  \textbf{ physical qualities of y and without change of the physical qualities of the retainer of y and without change of the mereotopology of y}.  
\end{definition}

\begin{elucidation}[Change of mereotopology]
	An entity changes its mereotopology if it gains or loses parts or the arrangements of its parts change. 
\end{elucidation}

For example,  the disposition of a cave to collapse may only cease to exist if its retainer (i.e., the surrounding rock) changes. However, the Geographic North Pole serves as a reference point for navigation not by disposition but by role, since it may lose this role without any alteration to its retainer (i.e., Earth). 
Further,  the disposition of `Hansel and Gretel' to scare children may only change if some of its parts are changed, e.g., by rewriting certain sections. However, its role as a running example of this paper was gained and may cease to exist without changing the mereotopology of the fairy tale. 

\section{Conclusion}
In this paper, we discussed some of the issues related to dispositions and roles of \gdcs in BFO 2020. We discussed two approaches to solving them. The first, as proposed by Barry Smith, is the introduction of defined classes. As we have seen, this approach has serious limitations. Further, we suggested a few minor changes to BFO, which would address the issue directly and introduce dispositions and roles of \gdcs. The suggested changes have the additional benefit of addressing existing problems of BFO 2020 with respect to roles and dispositions of sites and continuant fiat boundaries.

\begin{acknowledgments}
 I would like to thank Barry Smith, Alan Ruttenberg, and the anonymous reviewers for their insightful comments.  
 This work was funded by the Federal Ministry of Education and Research (BMBF) -- funding code 13XP5187E.
\end{acknowledgments}

\section*{Declaration on Generative AI}
   During the preparation of this work, the author used Grammarly in order to: Grammar and spelling check, paraphrase and reword. After using this tool/service, the author reviewed and edited the content as needed and takes full responsibility for the publication’s content.

\bibliography{bibliography}

\end{document}